\documentclass[twocolumn,letterpaper]{IEEEAerospaceCLS}  % only supports two-column, letterpaper format

\usepackage[]{graphicx}    % We use this package in this document
\newcommand{\ignore}[1]{}  % {} empty inside = %% comment

\usepackage{amsmath}
\usepackage{multirow}
\usepackage{caption}
\usepackage{subcaption}

\begin{document}
\title{Machine Learning Based Path Planning for Improved Rover Navigation (Pre-Print Version)}

\author{
Neil Abcouwer, Shreyansh Daftry, Tyler del Sesto, \\
Olivier Toupet, Masahiro Ono \\
Jet Propulsion Laboratory \\
California Institute of Technology\\
4800 Oak Grove Drive\\
Pasadena, CA 91109, USA\\ 
\textit{firstname.lastname}@jpl.nasa.gov
\and
Siddarth Venkatraman \\
Manipal Institute of Technology \\
Udupi - Karkala Rd, Eshwar Nagar, Manipal \\
Karnataka 576104, India \\
siddarth.venkatraman@learner.manipal.edu
\and
Ravi Lanka \\
Rakuten Inc. \\
19 Primrose Road, Bengaluru \\
Karnataka 560025, India \\
lanka.ravikiran@rakuten.com
\and
Jialin Song, Yisong Yue\\
Computing and Mathematical Sciences \\
California Institute of Technology \\
1200 E California Blvd \\
Pasadena, CA 91125, USA\\
\{jssong,yyue\}@caltech.edu
%%%% IMPORTANT: Use the correct copyright information--IEEE, Crown, or U.S. government. %%%%%
% JPL wants: If corresponding author is JPL, and there are non-JPL co-authors:  © YEAR. All rights reserved.
\thanks{\footnotesize \copyright 2020. All rights reserved. This work has been submitted to the IEEE for possible publication. Copyright may be transferred without notice, after which this version may no longer be accessible.}
%\thanks{\footnotesize \copyright 2020. All rights reserved.}  %FIXME apply this to submission?
%\thanks{\footnotesize 978-1-7281-7436-5/21/$\$31.00$ \copyright2021 IEEE}              % This creates the copyright info that is the correct 2021 data.
%\thanks{{U.S. Government work not protected by U.S. copyright}}         % Use this copyright notice only if you are employed by the U.S. Government.
%\thanks{{978-1-7281-7436-5/21/$\$31.00$ \copyright2021 Crown}}          % Use this copyright notice only if you are employed by a crown government (e.g., Canada, UK, Australia).
%\thanks{{978-1-7281-7436-5/21/$\$31.00$ \copyright2021 European Union}}    % Use this copyright notice is you are employed by the European Union.
}

% guidance on preprint: https://www.ieee.org/content/dam/ieee-org/ieee/web/org/pubs/author_version_faq.pdf 

\maketitle

\thispagestyle{plain}
\pagestyle{plain}

\maketitle

\thispagestyle{plain}
\pagestyle{plain}

\begin{abstract}
Enhanced AutoNav (ENav), the baseline surface navigation software for NASA’s Perseverance rover, sorts a list of candidate paths for the rover to traverse, then uses the Approximate Clearance Evaluation (ACE) algorithm to evaluate whether the most highly ranked paths are safe. ACE is crucial for maintaining the safety of the rover, but is computationally expensive. If the most promising candidates in the list of paths are all found to be infeasible, ENav must continue to search the list and run time-consuming ACE evaluations until a feasible path is found. In this paper, we present two heuristics that, given a terrain heightmap around the rover, produce cost estimates that more effectively rank the candidate paths before ACE evaluation. The first heuristic uses Sobel operators and convolution to incorporate the cost of traversing high-gradient terrain. The second heuristic uses a machine learning (ML) model to predict areas that will be deemed untraversable by ACE. We used physics simulations to collect training data for the ML model and to run Monte Carlo trials to quantify navigation performance across a variety of terrains with various slopes and rock distributions. Compared to ENav's baseline performance, integrating the heuristics can lead to a significant reduction in ACE evaluations and average computation time per planning cycle, increase path efficiency, and maintain or improve the rate of successful traverses. This strategy of targeting specific bottlenecks with ML while maintaining the original ACE safety checks provides an example of how ML can be infused into planetary science missions and other safety-critical software.
\end{abstract} 

\begin{figure}
\centering
\includegraphics[width=2.5in]{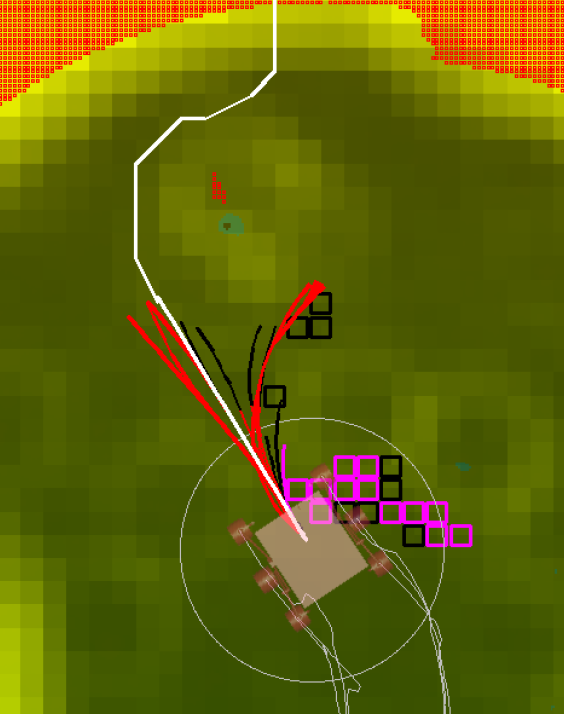}\\
\caption{\textbf{ A view of the ENav simulation environment. The green-yellow terrain shows how the Gradient Convolution heuristic, developed in this work, has assessed the cost of traversing the terrain (yellow regions are higher cost) and steers the rover toward safer regions. }}
\label{ENav_view}
\end{figure}

\tableofcontents

%%%%%%%%%%%%%%%%%%%%%%%%%%%%%%%%%%%%%%
\section{Introduction}
%%%%%%%%%%%%%%%%%%%%%%%%%%%%%%%%%%%%%%

The communications latency between Mars and Earth has driven Mars rover missions to develop significant surface autonomy. On-board planning \cite{chi2018embedding} generates schedules and models resource usage over time.
Autonomous target selection \cite{Franciseaan4582} is used to opportunistically gather science data on desirable specimens. Autonomous rover navigation \cite{mchenry2020navigation} allows rovers to traverse unknown terrain of increasing complexity and at greater rates without human drivers in the loop.

Future missions will demand even more capable autonomous mobility. The proposed Mars Sample Return mission's fetch rover will be more focused on fast and reliable mobility than gathering its own science data \cite{MUIRHEAD2020131}. The stated goals of the Mars Exploration Program Analysis Group (MEPAG) call for a greater spatial coverage of the Martian surface \cite{mepag2020}; this would necessitate a greater number of rovers and landers, which could not be operated without either a vastly increased amount of ground operators or more significant on-board autonomy. Missions to ocean worlds in the outer solar system are of great interest, but the increased communication delays may require rovers to operate autonomously for several earth days without human input\cite{sherwood2018program}.

The Mars 2020 mission \cite{williford2018nasa} and its Perseverance Rover will use the Enhanced Navigation (ENav) library \cite{toupet2020ENav} to plan paths on the Martian surface. ENav takes as input stereo imagery, maintains a 2.5D heightmap describing the terrain, and chooses the best maneuver to safely move the rover toward the global goal. ENav uses the Approximate Clearance Evaluation (ACE) algorithm \cite{otsu2020} to evaluate a sorted list of paths for safe traversal. Running the ACE algorithm on dozens of rover poses along hundreds of candidate rover paths represents a significant computational burden, especially if the list is sorted poorly and many paths fail the ACE check, which is more likely in complex and challenging terrain. 

In this paper, we describe ENav and its baseline performance (Section \ref{baseline}). We introduce two heuristics, one hand-designed and one machine-learned, that can be used to more effectively sort the list of candidate paths. Because the heightmap is represented as a two-dimensional array, a multitude of efficient techniques from computer vision can be used for analysis. We designed the Gradient Convolution heuristic to use convolution and Sobel operators \cite{sobel} to estimate the cost of terrain traversal (Section \ref{gradient_convolution}). We also used a data set of heightmaps and corresponding ACE evaluations to train a machine learning (ML) classifier to infer ACE values (Section \ref{learned_heuristic}). These two heuristics were integrated into ENav (Figure \ref{ENav_view}) and used to more effectively sort the rover paths before the ACE evaluation step.  By incorporating the ML classifier into the ranking process, but still checking paths for safety with ACE, we show how ML can be integrated into ENav without sacrificing safety requirements. We present our results for various experiments and describe how each heuristic affected the performance of ENav in Monte Carlo simulations across multiple terrains (Section \ref{experiments}). We show that integrating the heuristics improved path efficiency, greatly reduced ACE evaluations, computation time, and the likelihood of "overthinking" each planning cycle, and maintained or improved success rates compared to the baseline performance. Finally we discuss the implications and limitations of our results (Section \ref{conclusions}).

%%%%%%%%%%%%%%%%%%%%%%%%%%%%%%%%%%%%%%
\section{Baseline Navigation} \label{baseline}
%%%%%%%%%%%%%%%%%%%%%%%%%%%%%%%%%%%%%%

The ENav library is the core of the Mars 2020 autonomous rover navigation and constitutes the starting point and baseline performance level for this work.

At a high level, the ENav planning cycle involves following steps:

\subsection{Process Disparity}

As the rover drives, stereo cameras acquire images. Stereo correlation is used to generate a disparity image, which is then converted into a 3D point cloud. The point cloud is passed to ENav, which updates an internal 2.5D heightmap centered on the rover's position. Each point is added to the appropriate X-Y cell, and each cell with new points is assigned a height equal to the average height of the enclosed 3D points.

\subsection{Analyze Terrain} \label{AnalyzeTerrain}

Once the heightmap is updated, the terrain is analyzed and the results are saved in a costmap. Compared to the heightmap, the costmap is lower resolution, but covers a larger area. If a cell in the costmap corresponds to an area that was just imaged, the tilt angle and roughness (the average squared distance between a fitted plane and the points in the plane) are updated. The cost of the cell may be a weighted sum of the tilt, the roughness, and a minimum time needed to traverse a cell. The cell will be given an infinite cost if the tilt is extreme, plane-fitting fails, or if designated keep-in or keep-out zones (an area set by rover operators) are violated. Costmap cells representing areas that have not yet been imaged are given a finite cost value that is greater than that of known flat terrain, but more than that of known obstacles.

\subsection{Select Path}

With an updated costmap, ENav can select the next path for the rover to take to make progress towards the goal:

\subsubsection{Candidate Paths}

ENav considers a parameterized tree of candidate paths the rover can drive. The hardware of the Perseverance rover and its progenitors does not allow steering while driving. Thus the rover can move in fixed-curvature arcs (i.e. constant turning radius). Special cases of arcs include driving straight (zero curvature) and turning in place (infinite curvature). The first branch in the tree of paths generally consists of various turns (or no turning at all) and subsequent branches consist of fixed-length arcs of various curvatures.  The default tree of paths for our experiments is composed of 14 candidate turns-in-place (including the option of not turning), followed by 11 3-meter arcs of various curvatures, followed by another set of 11 3-meter arcs. This means the rover is planning about 6 meters ahead of its current position, and considering 1694 potential paths. 

\subsubsection{Initial Path Ranking} 

Each path in the tree is analyzed and assigned a total cost, which is a weighted sum of various cost factors. 
Firstly, each maneuver takes a certain amount of time to actuate; paths with less steering changes or no initial turn have a smaller cost. 
Secondly, the costmap (which was updated during the \textit{Analyze Terrain} step) is sampled at regular intervals along the maneuvers in each path. 
Finally, Dijkstra's algorithm \cite{dijkstra1959note} is used to estimate the minimum cost to travel from the endpoint of the considered path to the goal. After all paths have been assessed, the list of paths is sorted by cost. 

\subsubsection{ACE on Paths} 
Starting with the lowest-cost path, the ACE algorithm is applied at regular intervals along the path. Given the rover pose and heightmap, ACE finds the min and max heights within the potential footprint of each wheel, computes bounds on the rover attitude, suspension angles, and clearance between the rover belly pan and the terrain. If the bounds violate parameterized limits, ACE can return an infinite cost, rendering that path infeasible. Otherwise, a finite cost is returned based on proximity to the limits, which accumulates over the length of the path. Once a path with finite cost is found, ENav continues to analyze other candidate paths until reaching a threshold, and returns the lowest-cost ACE-validated path found.

\subsection{Actuation} 

Given the lowest-cost path, only the first maneuver (either the first meter of an arc or the first 30 degrees of a turn in place) of that path is actuated, stereo imagery is acquired and integrated, and the planning cycle repeats.

\subsection{Performance}

 This planning cycle starts when a disparity image is available for ENav, and ends when the solution is found. If the new maneuver is found before the rover finishes driving its current maneuver, the rover can immediately transition to the new maneuver. Otherwise, the rover must stop driving until a solution is found. Failing to find a solution before the rover stops is called "overthinking". Overthinking is undesirable, as it reduces the average traverse rate of the rover, increases wear and tear on the wheels and brakes, and increases mission risk overall. 

Overthinking is most likely to occur due to a poor ranking of candidate paths and excessive ACE evaluations taking significant computation time. In our experiments, we track ENav's overthink rate, defined as the percentage of planning cycles where the number of ACE evaluations exceeds 275. Each instance of ACE is estimated to take between 10 and 20 ms on the RAD750 computer used in the Perseverance rover, so this represents around 3 to 4 seconds of computation. In the theoretical scenario where the entire baseline tree of 1694 candidate paths were to be evaluated by ACE at 25 cm intervals, ACE would be called over 22 thousand times, taking over 3 minutes of computation time.

In benign terrain (terrain with little slope or few rocks, fully defined in section \ref{experiments}), the computational budget is rarely exceeded, but in complex terrain where unsafe obstacles are likely to occur between the rover and the goal, simulations predict an overthink rate of 20\%. The goal of this work is to reduce the planning computation time and increase the success rate and path efficiency in complex terrain. To achieve this, we added heuristics rank the potential rover paths more effectively with regards to the likelihood of passing ACE, reducing computation time and the overthink rate.

%%%%%%%%%%%%%%%%%%%%%%%%%%%%%%%%%%%%%%
\section{Gradient Convolution Heuristic} \label{gradient_convolution}
%%%%%%%%%%%%%%%%%%%%%%%%%%%%%%%%%%%%%%

\begin{figure}
     \centering
     \begin{subfigure}[t]{0.23\textwidth}
         \includegraphics[width=\textwidth]{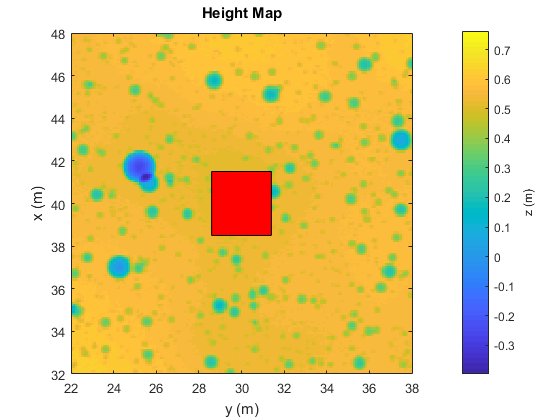}
         \caption{Height Map}
         \label{fig:heightmap}
     \end{subfigure}
     \hfill
     \begin{subfigure}[t]{0.23\textwidth}
         \includegraphics[width=\textwidth]{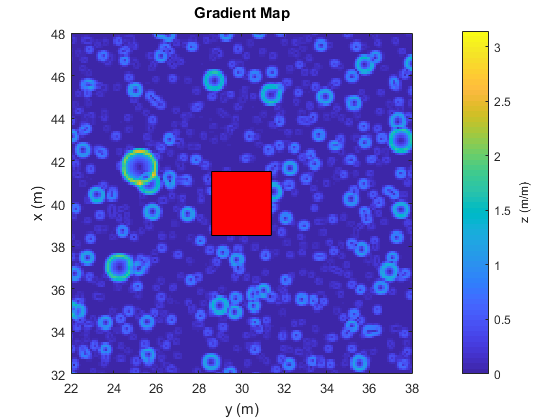}
         \caption{Gradient Map}
         \label{fig:gradientmap}
     \end{subfigure}
\\
\vfill
        \begin{subfigure}[t]{0.23\textwidth}
         \includegraphics[width=\textwidth]{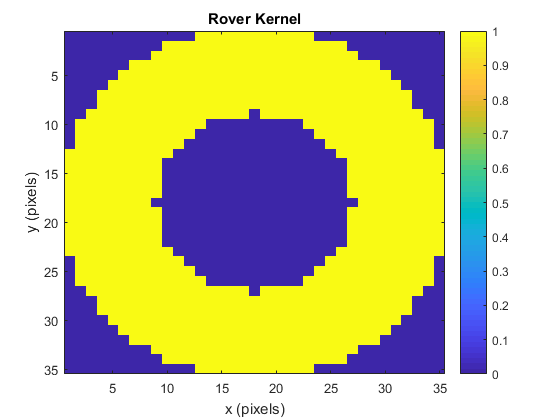}
         \caption{Rover Kernel}
         \label{fig:roverkernel}
     \end{subfigure}
\hfill
        \begin{subfigure}[t]{0.23\textwidth}
         \includegraphics[width=\textwidth]{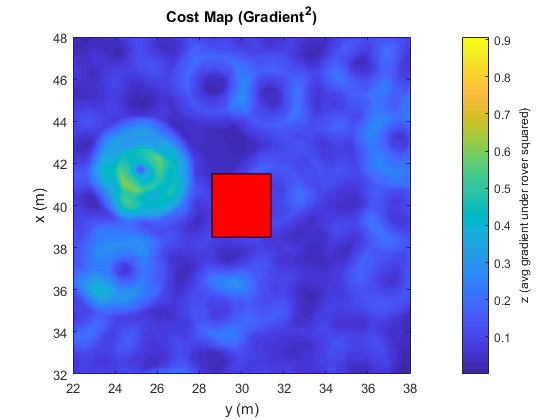}
         \caption{Cost Map}
         \label{fig:costmap}
    \end{subfigure}

  \caption{\textbf{Given a height map (a), Sobel operators can be used to create a gradient or gradient squared map (b). The gradient map can be convolved with a kernel representing the orientation-agnostic footprint of the rover (c) to form the gradient cost map (d), which estimates the cost to traverse a location within the map. }}
  \label{fig:maps}
\end{figure}

In order to more effectively sort ENav's list of candidate rover paths such that the most highly ranked paths are more likely to pass ACE, we designed the Gradient Convolution heuristic to estimate costs that would be correlated with ACE evaluations, but with a lower computation time. 

The ACE algorithm considers how uneven the terrain is under the rover wheels. The Gradient Convolution heuristic was designed to assess terrain roughness at the points where the rover wheels might be. Given a heightmap, the heuristic is computed as follows: 

\begin{enumerate}
\item Convolve the heightmap (Figure \ref{fig:heightmap}) with normalized 3x3 Sobel operators to find the local x and y gradient:

\begin{equation}
{\bf G_x } = \frac{1}{2r} \frac{1}{4} \begin{bmatrix} +1 & 0 & -1 \\ +2 & 0 & -2 \\ +1 & 0 & -1\end{bmatrix} * {\bf A}
\end{equation}

\begin{equation}
{\bf G_y} = \frac{1}{2r} \frac{1}{4}  \begin{bmatrix} +1 & +2 & +1 \\ 0 & 0 & 0 \\ -1 & -2 & -1\end{bmatrix} * {\bf A}
\end{equation}

where ${\bf A} $ is the heightmap, $r$ is the width of one square cell in the heightmap, and $*$ is the convolution operator.

\item Find the squared gradient magnitude map (Figure \ref{fig:gradientmap}) as:

\begin{equation}
{\bf G_{sq}} = {\bf G_x} \circ {\bf G_x} + {\bf G_y} \circ {\bf G_y}
\end{equation}

where $\circ$ is element-wise multiplication. Using the squared gradient magnitude is less computationally expensive (as no square root is required) and creates a quadratic cost, penalizing extreme values.

\item Convolve the squared gradient map with a kernel representing the heading-agnostic footprint of the rover (Figure \ref{fig:roverkernel}). The footprint is an annulus, with an outer radius corresponding to the furthest extent of any wheel and an inner radius corresponding the closest any wheel gets to the pivot point. If elements of the kernel are within the annulus, they have a value of 1, otherwise 0.  Normalize by dividing by the sum of the non-zero elements in the footprint kernel, and multiply by a parameterized cost factor.  

\begin{equation}
{\bf G_c} =   \frac{k}{\sum {\bf R}_i}  {\bf R} * {\bf G_{sq}}    
\end{equation}

where ${\bf R}$  is the rover kernel, and $k$ is the cost factor. This provides the gradient convolution cost map (Figure \ref{fig:costmap}).

\end{enumerate}

Convolution and the Sobel operators are very commonly used in computer vision contexts. As computer vision becomes increasingly important for rover navigation, and dedicated vision-processing computers are added to Mars missions \cite{mchenry2020navigation}, the likelihood of these methods being computed quickly and efficiently is high.

\subsection{Integration}

During the Analyze Terrain step of the ENav planning cycle (discussed in Section \ref{AnalyzeTerrain}), the gradient convolution cost is calculated for each cell in the costmap and included in the overall cost of the cell. This causes the heuristic to contribute not only to the cost of terrain traversal along each path, but also to the Dijkstra-based estimate of cost from the end of each path to the overall goal.

%%%%%%%%%%%%%%%%%%%%%%%%%%%%%%%%%%%%%%
\section{Learned Heuristic} \label{learned_heuristic}

\begin{figure*}[h]
\centering
\includegraphics[width=6.5in]{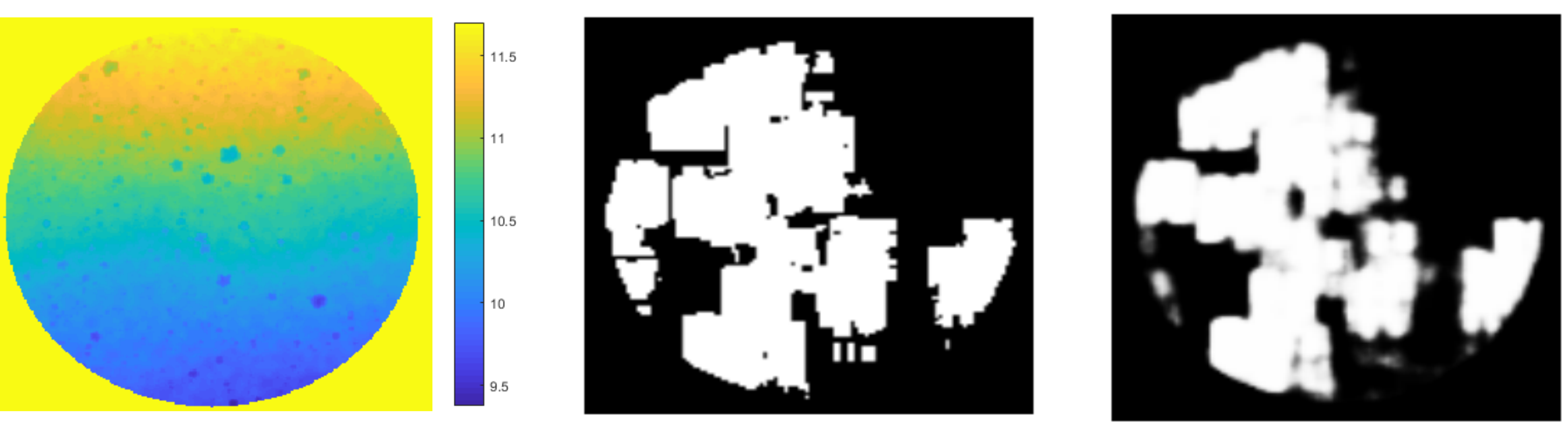}\\
\caption{\textbf{An example of a Learned Heuristic. Sets of terrain heightmaps (left) and maps generated by the ACE algorithm (center) were used to train a neural network to generate an inferred ACE probability map (right). }}
\label{LH}
\end{figure*}

In order to find a more accurate heuristic for ACE cost, we trained a model to predict ACE values based on heightmap data (Figure \ref{LH}). In contrast to the Gradient Convolution heuristic, which was hand-coded by domain experts, the learned heuristic is automatically encoded using a data-driven framework. More specifically, we developed a deep convolutional neural network (DCNN) based model that can directly predict the outcome of the ACE algorithm for a given terrain heightmap. Using this prediction, ENav can more optimally sort its initial list of potential paths and hence reduce the average number of ACE evaluations required until finding a safe path. 

\subsection{Model Architecture and Training}
We formulate this problem as a supervised-learning based classification. Our DCNN model is based on a modified encoder-decoder style U-Net architecture \cite{ronneberger2015u}, and implemented using the Tensorflow framework \cite{abadi2016tensorflow}. The encoder consists of a series of convolutional layers that down-samples the input to a low-dimensional feature map, and a decoder that consists of up-sampling layers with convolutions that then take this feature map and increase their resolution to that of the original input. U-Net also has a series of residual connections from the encoder to the decoder feature maps that helps restore the high-resolution details lost during down-sampling and also prevents vanishing gradients during training.

The input to our model is a heightmap and the output is an ACE map, such that the value for each pixel in the ACE map corresponds to the expected ACE cost for the corresponding terrain parameters. However, ACE cost depends not only on the terrain but also on the rover heading. We encode the rover heading as part of the learning problem itself by extending the output to have a multi-channel representation such that each channel represents a cardinal heading angle for the rover. In our experiments, we have found a discretization of 8 heading angles (at 45 degree intervals) to be sufficient. Sigmoid activation is applied to each channel to give a value in the range [0, 1] corresponding to the probability of a cell being infinite ACE cost or not.

Training data was gathered by running a Monte Carlo simulation of the baseline ENav algorithm on 1500 terrains, randomly sampling 8 heightmaps from each trial. For each cell in each sampled heightmap, the ACE algorithm was run with the eight fixed rover heading values, resulting in an “ACEmap” where each cell has eight heading-specific values. Of the 12000 total heightmap, ACE map pairs, 9500 were used as a training set, and 2500 were used as the validation set. The learned heuristic model achieved 97.8\% training accuracy and 95.3\% validation accuracy. 

Note the result of this prediction is a probability of ACE returning a safety violation, which is different from the output of the ACE algorithm itself. ACE can return finite or infinite costs, where finite costs represent how close the rover is to safety violations, and infinite costs represent safety violations. Predicting the actual ACE costs is a dual problem of classification and regression, and is more difficult that segmentation alone. Efforts to predict these values have not yet yielded results. 

\subsection{Integration}

To integrate the model inference into the ENav algorithm, the heightmaps are passed to a TensorFlow process, which returns an ACE prediction map to ENav. When ENav does the Initial Path Ranking step, the ACE map is sampled at regular intervals along the paths, proportionately weighting the predictions for the two fixed headings closest to the current heading. The ACE predction is multiplied by a cost factor, and adds to the accumulated cost of each path. The average ACE map value also contributes to the costmap and is used for the Dijkstra-based estimate of cost from the end of each oath to the overall goal. 

Importantly, the sorted list of paths is still evaluated by the “true” ACE algorithm. This approach allows us to leverage the benefits of ML while maintaining the same safety guarantees.

%%%%%%%%%%%%%%%%%%%%%%%%%%%%%%%%%%%%%%

%%%%%%%%%%%%%%%%%%%%%%%%%%%%%%%%%%%%%%
\section{Experiments} \label{experiments}
%%%%%%%%%%%%%%%%%%%%%%%%%%%%%%%%%%%%%%

\begin{figure}
\centering
\includegraphics[width=3.25in]{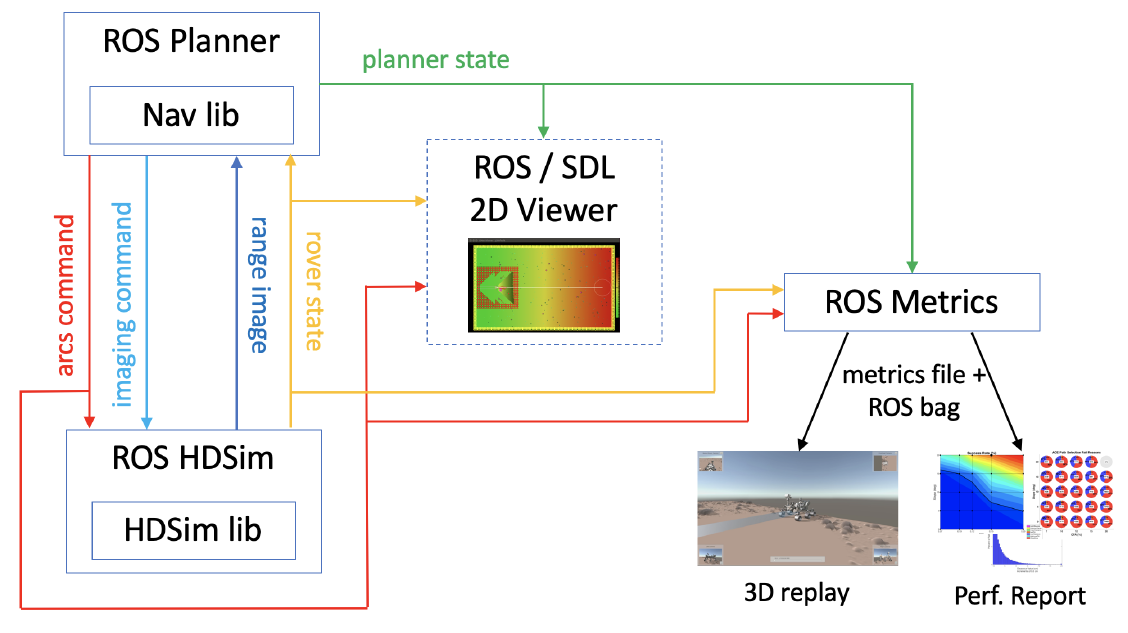}\\
\caption{\textbf{ROS software architecture for testing ENav against the HDSim physics simulator}}
\label{simulator_architecture}
\end{figure}

Described at length in \cite{toupet2020ENav}, a Monte Carlo simulation environment was built for testing ENav and Mars 2020 navigation (Figure \ref{simulator_architecture}). The Robotics Operating System (ROS) \cite{quigley2009ros} was used for inter-process communication. One ROS node wraps ENav, and another wraps the HyperDrive Simulator (HDSim), which simulates rover motion, terrain settling and slipping, and disparity images for JPL rover missions. Simulated terrains, representing various slopes and rock densities, are loaded into HDSim. Rock densities are classified by the cumulative fractional area covered by rocks (CFA) \cite{golombek1997size}. 

The rover starts at one side of the map and ENav is given a global goal 80m away at the other end of the map. The trial is run until the rover either successfully reaches the goal, or until a failure condition is found, such as when no feasible path can be found, when the safety limits of the rover are violated, or when the duration of the trial exceeds a time limit. 

We leverage this existing simulation setup to run our experiments. In the case of simulations employing the machine learning model, we use an additional ROS node running TensorFlow, which receives heightmaps and publishes ACE estimates. Because HDSim currently only runs on 32-bit systems, and TensorFlow only runs on 64-bit systems, we use a ROS multi-master system to communicate between two computers. 

The discussion of the experiments notes potential gains in computation time. These gains are theoretical and predicated on the notion that each call of the ACE algorithm takes 10 to 20 ms on the RAD750 flight processor, these calls make up a very large portion of ENav's computation time, and reduction of these calls translates to a gain in computation time. The use of non-flight-like computers, the need for multi-master ROS, and lack of optimization in the added algorithms for this research prevent meaningful comparisons of wall clock time.

\begin{figure*}[h]
\centering
\includegraphics[width=7in]{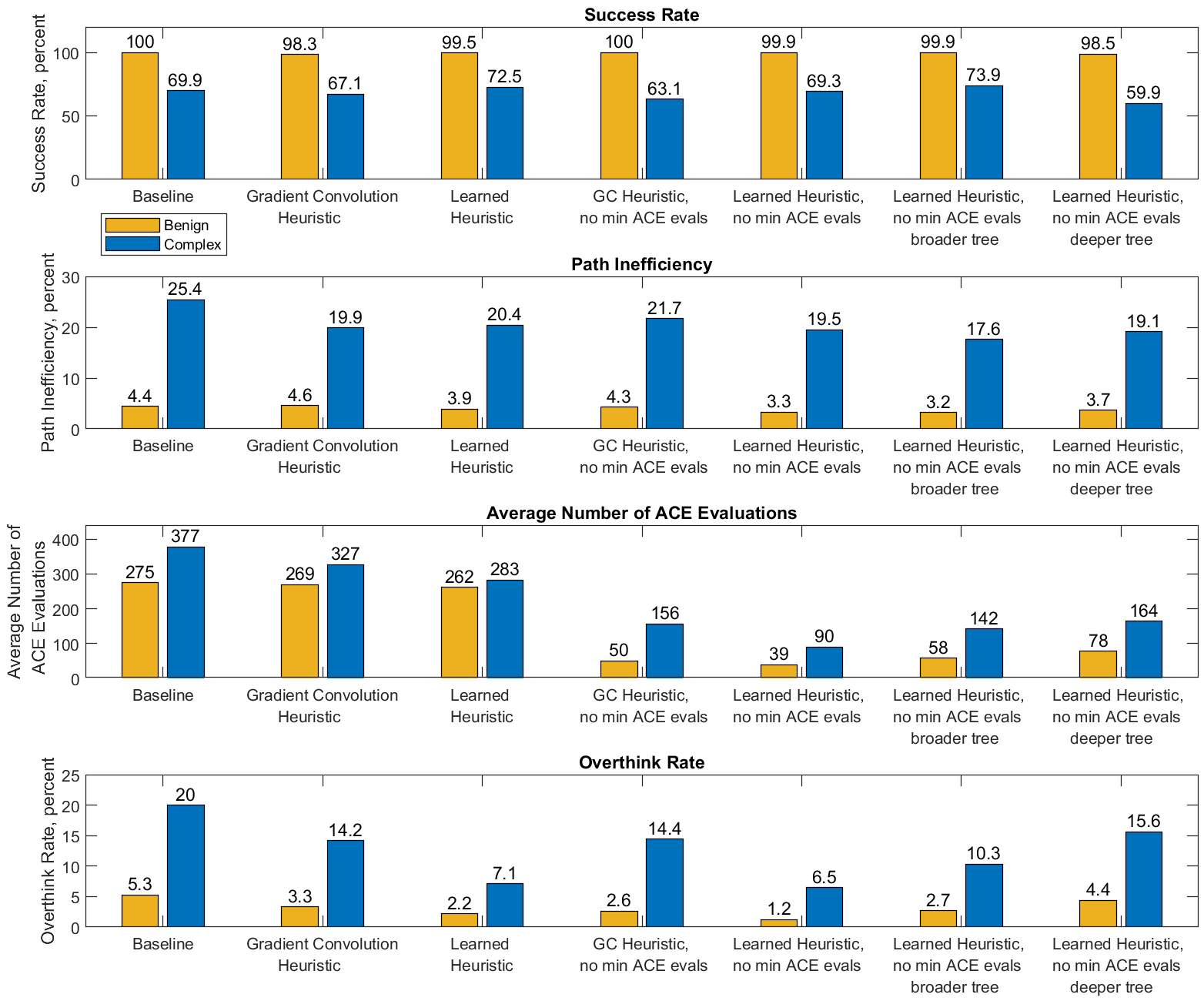}\\
\caption{\textbf{A summary of key rover path planning performance metrics across various experiments. Incorporating heuristics produced more efficient paths, reduced the number of costly ACE evaluations, and maintained or slightly increased the rate of successfully reaching the goal.}}
\label{ExperimentsFig}
\end{figure*}

The results of the following experiments are summarized in Figure \ref{ExperimentsFig}. Each simulation tracks results on two subsets of terrains: \textbf{Benign} terrains have a CFA value of 7\% or less, and a slope of 15$^{\circ}$ or less. \textbf{Complex} terrains have greater slope or CFA. Each performance metric is calculated as a weighted average across each subset of terrains (Benign or Complex), with terrains of greater complexity being less likely to occur on Mars, and accordingly given less weight. We are more concerned with tracking performance on complex terrain. Within each Monte Carlo simulation, 780 trials are run on complex terrain, so $n=780$ will be used when calculating 95\% confidence margins of error (MOE).

The following performance metrics are tracked:

\begin{itemize}
\item \textbf{Success Rate} is the percentage of trials for each terrain that result in the rover reaching the global goal without timing out, reaching a point with no feasible paths, or violating safety constraints. Higher values are better.
 \item \textbf{Average Path Inefficiency} is defined as the average length of the path taken by the rover divided by the Euclidean distance from the start to the goal, minus 1, expressed as a percentage. For example, if the rover goal was 100 m away, and the rover needed to travel a circuitous route with a length of 125 m to avoid obstacles and reach the goal, the path inefficiency was 25\%. Lower values are better.
\item \textbf{Average ACE Evaluations} is the average number of ACE evaluations conducted per planning cycle. Lower values are better. Each evaluation takes an estimated 10 to 20 ms on a RAD750 processor like that on Perseverance, so a conversion to average cycle time can easily be made. 
\item \textbf{Overthink Rate} is the average percent of ENav planning cycles that required more ACE evaluations than a threshold value (275 by default), which indicates that the highest-ranked candidate paths were all deemed unsafe by ACE, and therefore the initial ranking of paths was unsuitable. When the number of ACE evaluations exceeds the threshold, it indicates that ENav is "overthinking" and the rover may need to stop driving until a solution is found. Lower values are better. 
\end{itemize}

\subsection{Experiment 1: The effects of heuristics}

The first experiment tested whether adding the designed and learned heuristics to the baseline ENav software yields gains in the tracked performance metrics. Monte Carlo simulations were run with the same terrains and parameters as the baseline simulation. Note that the terrains used for these experiments are a separate set from the set of terrains used to train the learned model, to prevent unrepresentative performance due to over-fitting.

The simulation with the gradient convolution heuristic showed mixed results (Figure \ref{ExperimentsFig}). Benign terrain performance was fairly similar to the baseline. Success rate dropped slightly, but the overthink rate decreased. Complex performance showed more obvious changes. Success rate slightly worsened, from 69.9\% to 67.1\%, but other metrics were promising. There were significant decreases in path inefficiency, from 25.4\% to 19.9\% (MOE: 2.8\%). In other words, given a goal 100 m away, the new method would be expected to drive a 120 m path, compared to 125 m for the baseline, saving time and energy. The ACE metrics also improved, with the overthink rate reducing significantly, from 20.0\% to 14.2\% (MOE: 2.5\%). In summary, the designed heuristic did show promise for improving ENav performance in complex terrains.

More impressively, the simulation with the learned heuristic showed improvements across almost every metric. Complex success rate improved slightly, from 69.9\% to 72.5\% (within the MOE). Path inefficiency significantly improved, especially for complex terrains, going from 25.4\% to 20.4\%. The number of ACE evaluations also reduced, especially in terms of the overthink rate, which plummeted to 7.1\% compared to 20.0\% for the baseline. This would result in far fewer cases of the rover needing to stop driving before the next path can be found. 

We asserted that ML could be added without sacrificing the safety guarantee of the ACE algorithm, and this assertion holds. No trial failures were caused by violated safety constraints; all failures are due to either timeouts or failures to find paths to the goal. This was true for all experiments.

This experiment answers the most fundamental question of this work: can heuristics, designed or learned, improve the performance of the baseline rover navigation algorithm? The answer is yes. Heuristics can more effectively rank the set of candidate paths, reduce the average computation time needed to find a safe path, choose maneuvers that increase path efficiency, and increase the likelihood of successfully reaching the goal, while maintaining rover safety.

\subsection{Experiment 2: No minimum evaluations if solution found}

As discussed in Section \ref{baseline}, by default when a path that passes ACE is found, ENav will continue to evaluate paths up to a parameterized limit. This helps if the first safe path has a poor (but finite) ACE cost; a path with a lower cost relative to ACE evaluations can be found. 

But if these heuristics rank candidate paths more effectively, then this continued evaluation may no longer be necessary and computational resources might be better used elsewhere. This second experiment quantifies how much computation time can be saved, and a what cost to other metrics. We repeated the simulation with each heuristic, this time setting up ENav parameters such that the first ACE-feasible path from the sorted list is chosen for the next maneuver.
 
The simulation with the designed heuristic showed similar results to Experiment 1, but with a vast reduction in the number of ACE evaluations. The average number of ACE evaluations in complex terrain was only 156, compared to 377 for the baseline. On the Perseverance RAD750 processor, this would translate to an average ACE computation time of around 1.9 seconds, rather than 4.5 seconds for the baseline. However the complex success rate reduced to 63.1\%. 

The simulation with the ML heuristic showed promising results. Success rates remained similar to the baseline, but the complex success rate of 69.3\% does represent a small decrease from learned heuristic success rate (72.5\%) in Experiment 1. Path efficiencies improved even more than Experiment 1, but the biggest improvement was in the average number of ACE evaluations. In benign terrain, the average number of ACE evaluations went from 275 to 39. In Complex terrain, ACE Evaluations went from 377 to 90. For a RAD750 processor, this would represent an improvement from about 4.5 seconds to 1.1 seconds, saving 3.4 seconds of computation time per planning cycle. 

This experiment showed that the inclusion of learned heuristics needed less than one fourth the number of ACE evaluations, and therefore significantly reduced time, to achieve similar success rates as the baseline software.

\subsection{Experiment 3: Using a broader tree of paths}

Experiment 2 showed that using the ML heuristic could significantly reduce computation time in the ENav planning cycle. Reduction of computation time is only useful if that freed computational budget is used for methods that might increase the success rate. One potential approach is to increase the number of candidate paths, which can increase the probability of finding a safe path, or provide a wider range from which to choose a more efficient path through the terrain. 

As discussed in Section \ref{baseline}, the baseline tree of paths has 14 potential tuns-in-place (including no turn) at depth 1, followed by two levels of 11 potential arcs, for a total of 1694 paths. For Experiment 3, the possible choices at each level of the tree was increased by 4, giving 4050 possible paths, more densely covering the terrain. This means that analyzing all paths could take around 2-3 times as much time, if not effectively sorted by the heuristics.

Experiment 3 was conducted with the machine learning model in the loop, no minimum number of ACE evaluations, and the broader tree. The experiment showed the best success rates and path efficiencies so far, improving complex success rate from 69.9\% to 73.9\% (MOE: 3.1\%) and inefficiency from 25.4\% to 17.6\%. While timing results showed more average ACE evaluations and a higher chance of overthinking than Experiment 2, these metrics were still significantly superior to the baseline. However, there would be an increased amount of time required to perform the initial ranking of the larger set of paths, which is not captured by our current experimental setup. 

Because prior experiments showed that the learned heuristic was superior to the designed heuristic, further experiments with the designed heuristic were not conducted. 

\subsection{Experiment 4: Using a deeper tree of paths}

Experiment 3 showed how assessment of a broader tree of paths becomes viable and shows high performance with the addition of the learned heuristic. The next experiment tests whether a deeper tree can also yield promising results. We add an additional set of 11 arcs to the previous leaf nodes in the tree, extending the planning horizon of the rover to nine meters ahead. 

Merely adding another depth of 11 arcs to the baseline tree ($14 \textrm{ turns} \times 11 \textrm{ arcs} \times 11 \textrm{ arcs} = 1694 \textrm{ paths} $), would cause the time required to analyze the tree to multiply by 11. Instead, we evaluate the tree up to the depth of two arcs, prune away all but the lowest cost paths within each set of 11 depth-3 arcs, making a tree of ($14 \textrm{ turns} \times 11 \textrm{ arcs} \times 1 \textrm{ arc} = $) 154 paths, then extending and evaluating 11 arcs from the remaining paths. This combats the branching factor, keeping the number of possible paths the same and roughly doubling the amount of computation required to analyze the paths, in line with the twice-as-large tree from Experiment 3. 

The results from this approach were poor. The success rate in complex terrain tumbled to 59.9\%, and no other success metrics improved relative to Experiments 2 or 3. Additional complexity may have caused trial timeouts to be more likely. These losses were not offset by gains in planning; stereo estimation of terrain nine meters away may be too inaccurate or extending the paths to nine meters may not be useful when slip is accumulated over that distance. Better pruning strategies may yield better results in future experiments.

%%%%%%%%%%%%%%%%%%%%%%%%%%%%%%%%%%%%%%
\section{Conclusions}  \label{conclusions}
%%%%%%%%%%%%%%%%%%%%%%%%%%%%%%%%%%%%%%

Our experiments show that heuristics for terrain safety can improve the performance of the baseline ENav software. 

All the heuristic experiments showed significantly increased path efficiency in complex terrain. This may be because the ACE algorithm is likely to find narrow safe paths that disappear once slip occurs, causing backtracking or tight turns. In contrast, the heuristics form a smoother estimate and steer the rover away from such problematic areas.

The experiments with no minimum number of evaluations per planning cycle showed that the computation time associated with ACE evaluations can be greatly reduced with only small decreases in success rate. The success rate can be increased by using the freed computation time to evaluate a broader set of paths, yielding results superior to the baseline in every performance metric. Other methods might leverage the computation for even better results.

We found that the ML model outperformed the hand-designed heuristic in most of the performance metrics in most of the experiments, but the designed heuristic showed potential improvements compared to the baseline ENav performance. The designed heuristic could be considered for further testing in representative hardware and for architectures where ML inference can not be deployed. 

An important caveat to our claims of improved performance is that gains in computation time are measured purely in terms of reduced ACE evaluations. If the time savings are completely consumed by increased time used to calculate the heuristics, these claims are no longer valid. We hope for opportunities to more accurately test the performance in a flight-like computing environment.

More broadly, these results offer an example of how ML can be used to improve the performance of a safety-critical flight system without sacrificing safety guarantees. While the ACE-estimating model is a black box, the result is used only for ranking, not for safety-critical decisions. Any path returned from ENav has passed a safety evaluation by the true ACE algorithm. This method could be applied to various critical robotic or spaceflight tasks, such as scheduling or kinematics, where the range of possible solutions is vast and calls for an inference-based approach, but the result can be evaluated by a known traditional algorithm that can be verified and validated.

%%%%%%%%%%%%%%%%%%%%%%%%%%%%%%%%%%%%%%%%%%%%%%%%%%%%%%%%%%%%%%%%%%%%%%%%%%%%%%%%%%%%%%%%%%%%%%%%%%%%%%
\acknowledgments
The research described in this paper was performed at the Jet Propulsion Laboratory, California Institute of Technology, under a contract with the National Aeronautics and Space Administration (80NM0018D0004). The authors would like to thank the JPL Research and Technology Development (R\&TD) program for supporting this research.

%%%%%%%%%%%%%%%%%%%%%%%%%%%%%%%%%%%%%%%%%%%%%%%%%%%%%%%%%%%%%%%%%%%%%%%%%%%%%%%%%%%%%%%%%%%%%%%%%%%%%%
\bibliographystyle{IEEEtran}
\bibliography{IEEEabrv,abcouwerMlnavIEEEA2021_preprint}

%%%%%%%%%%%%%%%%%%%%%%%%%%%%%%%%%%%%%%%%%%%%%%%%%%%%%%%%%%%%%%%%%%%%%%%%%%%%%%%%%%%%%%%%%%%%%%%%%%%%%%
\thebiography
%% This biostyle allows you to insert your photo size 1in X 1.25in
\begin{biographywithpic}
{Neil Abcouwer}{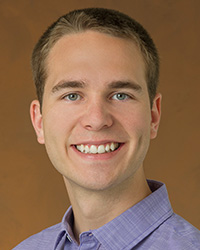}
received his M.S. in Robotics from Carnegie Mellon University. He is a Robotics Electrical Engineer at the Jet Propulsion Laboratory. He has worked on various robotic software projects, including mobility, pointing, pose estimation, and compression modules for the Perseverance Mars rover. 
\end{biographywithpic} 

\begin{biographywithpic}
{Shreyansh Daftry}{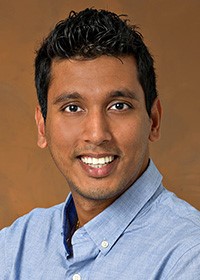}
received his M.S. in Robotics from Carnegie Mellon University. He is a Robotics Technologist at Jet Propulsion Laboratory. At JPL, he has worked on mission formulation for Mars Sample Return, and technology development for autonomous navigation of ground, airborne and subterranean robots.
\end{biographywithpic}

\begin{biographywithpic}
{Siddarth Venkatraman}{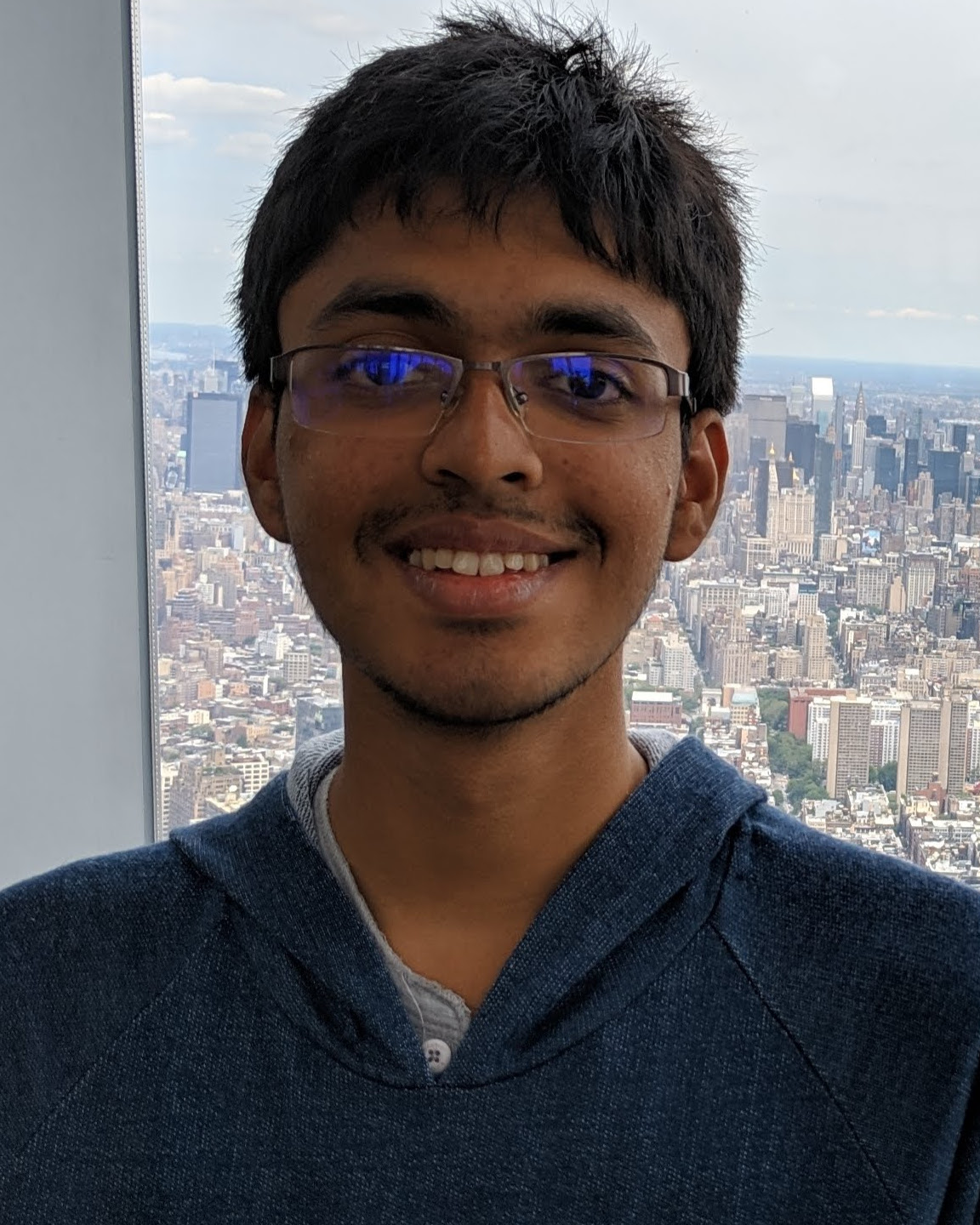}
is working toward his B.E. in Computer Science and Computational Mathematics at Manipal Institute of Technology. He was an Intern in the Perception Systems group at the Jet Propulsion Laboratory. He has worked on perception and control implementations with Machine Learning for various robotics projects, including a semi-autonomous car. 
\end{biographywithpic}

\begin{biographywithpic}
{Tyler del Sesto}{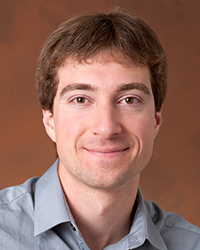}
received his M.S. in Mechanical Engineering from Carnegie Mellon University. 
He is a Robotic Systems Engineer at the Jet Propulsion Laboratory. He has worked robotic operations for the Curiosity rover as well as software and systems level testing of mobility and autonav for the Perseverance rover.
\end{biographywithpic}

\begin{biographywithpic}
{Olivier Toupet}{otoupet}
received his M.S. in Aeronautics and Astronautics from the Massachusetts Institute of Technology. He is currently the supervisor of the Robotic Aerial Mobility Group at the Jet Propulsion Laboratory. His research also includes the development of novel robotic technologies for Mars rovers, such as the path planner for the Perseverance rover mentioned in this paper, and the traction control algorithm for the Curiosity rover. 
\end{biographywithpic}

\begin{biographywithpic}
{Ravi Lanka}{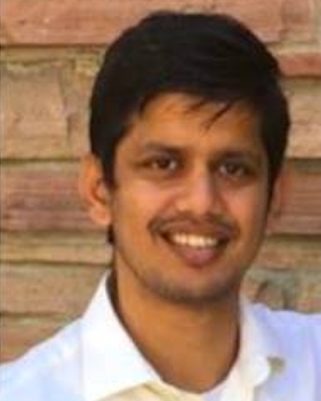}
received his M.S. in Computer Science from New York University. 
He is currently a Research Scientist at Rakuten, India. He was previously a Research Technologist in the Machine Learning and Instrument Autonomy group at the Jet Propulsion Laboratory. 
His research interests lie in developing scalable approaches to combinatorial optimization problems, by leveraging the interplay between machine learning and optimization. 
\end{biographywithpic}

\begin{biographywithpic}
{Jialin Song}{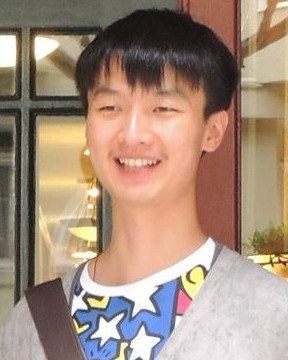}
received his B.S. in Computer Science and Mathematics from the University of Toronto. He is currently a Ph.D. candidate at California Institute of Technology. He has worked on various projects on using machine learning to speed up solvers for optimization problems such as planning and structural optimization.
\end{biographywithpic}

\begin{biographywithpic}
{Yisong Yue}{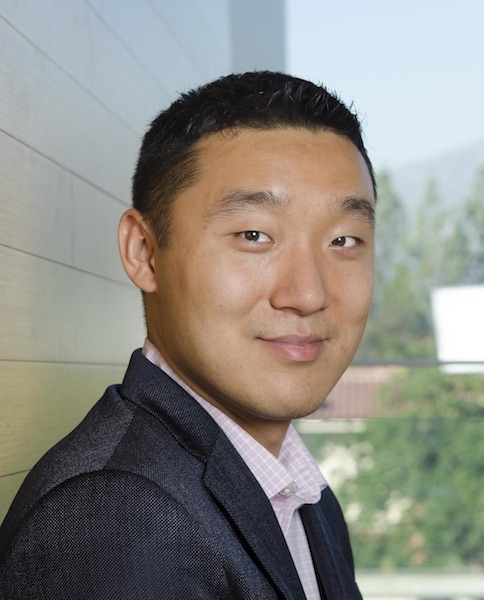} 
received his Ph.D. in Computer Science from Cornell University. He is a Professor of Computing and Mathematical Sciences at Caltech.  His research interests span both the theory and application of machine learning.  His research has been applied to recommender systems, robotics, planning, automation, and experiment design in science.
\end{biographywithpic}

\begin{biographywithpic}
{Masahiro Ono}{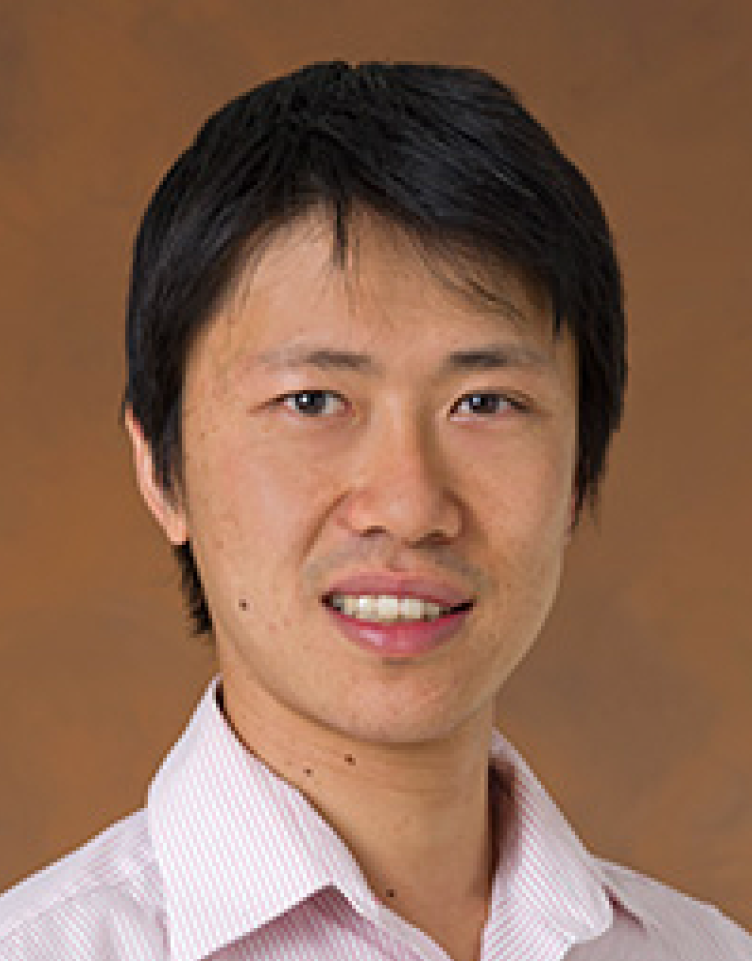}
received his Ph.D. in Aeronautics and Astronautics from the Massachusetts Institute of Technology. He is a Research Technologist at the Jet Propulsion Laboratory. His broad interest is centered around the application of robotic autonomy
to space exploration, with an emphasis on machine learning applications to perception, data interpretation, and decision making. 
\end{biographywithpic}

\end{document}